\documentclass[11pt,a4paper]{article}
\usepackage[hyperref]{acl2021}
\usepackage{times}
\usepackage{latexsym}
\usepackage{amssymb}
\usepackage{booktabs}
\usepackage{rotating}
\usepackage{xcolor}
\usepackage{hyperref}
\usepackage{algorithm}
\usepackage{algpseudocode}

\usepackage{microtype}

\aclfinalcopy 

\title{Modality and Negation in Event Extraction}

\author{Sander Bijl de Vroe*, Liane Guillou*, Miloš Stanojević, Nick M{\normalsize\raisebox{0.3ex}{c}}Kenna, Mark Steedman \\
School of Informatics, University of Edinburgh\\
\texttt{\{sbdv, liane.guillou, m.stanojevic, nick.mckenna\}@ed.ac.uk} \\
\texttt{steedman@inf.ed.ac.uk}}

\date{}

\begin{document}
\maketitle

\begingroup\renewcommand\thefootnote{*}
\footnotetext{The first two authors contributed equally to this work}
\endgroup

\begin{abstract}
Language provides speakers with a rich system of modality for expressing thoughts about events, without being committed to their actual occurrence. Modality is commonly used in the political news domain, where both actual and possible courses of events are discussed. NLP systems struggle with these semantic phenomena, often incorrectly extracting events which did not happen, which can lead to issues in downstream applications. We present an open-domain, lexicon-based event extraction system that captures various types of modality. This information is valuable for Question Answering, Knowledge Graph construction and Fact-checking tasks, and our evaluation shows that the system is sufficiently strong to be used in downstream applications.
\end{abstract}

\section{Introduction}

Linguistic modality is frequently used in natural language to express uncertainty with respect to events and states. Downstream NLP tasks that depend on knowing whether an event actually occurred, such as Knowledge Graph construction, Fact-checking, Question Answering and Entailment Graph construction, can benefit from understanding modality. Such information is crucial in the medical domain, for instance, where it facilitates more accurate Information Extraction and search for radiology reports \citep{wu2011evaluation, peng2018negbio}. Similarly, if we pose a question in the socio-political domain, such as \textit{Did the protesters attack the police?}, our answer will be different depending on the evidence that the system has observed: \textit{Protesters attacked the police} [yes] or \textit{Protesters are unlikely to have attacked the police} [uncertain]\footnote{Assuming trustworthy source text}.

These challenges are exacerbated by the prevalence of the phenomenon. In a multi-domain uncertainty corpus \citep{szarvas2012cross}, sentences containing uncertainty cues are significantly more common in newswire text (18\%) compared to encyclopedic text (13\%). Modality is also frequently observed in editorials \citep{bonyadi2011linguistic}. We show that within the news genre, modality is common in the politics and sports domains, where experts often make predictions and state their opinions on the possible outcomes of events such as elections or sports matches, and analyse alternative outcomes where situations unfold differently. 

We present \textsc{MoNTEE}\footnote{\href{https://gitlab.com/lianeg/montee}{https://gitlab.com/lianeg/montee}}, an open-domain system for \textbf{Mo}dality and \textbf{N}egation \textbf{T}agging in \textbf{E}vent \textbf{E}xtraction. Tagging these phenomena allows us to distinguish between events that took place (e.g. \textit{Protesters attacked the police}), those that did not take place (\textit{Had protesters attacked the police...}), or are uncertain at the time that a document is written (\textit{Protesters may have attacked the police}). 

The extracted relations include a predicate and one or two arguments, for example: \textbf{Protesters}-\textit{attack}-\textbf{police} (from the sentence \textit{Protesters attacked the police}). The predicates are analysed according to the following semantic phenomena: negation, lexical negation, modal operators, conditionality, counterfactuality and propositional attitude. See Table~\ref{table:modality_categories} for examples of each category.

\begin{table}
\small
\centering
\begin{tabular}{@{}ll@{}}
    \hline
    \textbf{Category} & \textbf{Example} \\
    \hline
    $\varnothing$ & Protesters attacked the police \\
    Negation & Protesters did \textbf{not} attack the police \\
    Lexical negation & Protesters \textbf{refrained} from\\
    & attacking the police \\
    Modal operator & Protesters \textbf{may} have attacked the police \\
    Conditional & \textbf{If} protesters attack the police... \\
    Counterfactual & \textbf{Had} protesters attacked the police... \\
    Propositional & Journalists \textbf{said} that \\
    {    }attitude & protesters attacked the police \\
    \hline
\end{tabular}
\caption{Modality and negation categories}
\label{table:modality_categories}
\end{table}

We contribute a lexicon of words and phrases that trigger modality, a parser that extracts and tags open-domain event relations for modality (along with an intrinsic evaluation), and a corpus study focusing on the politics domain of a large corpus of news text.

\section{Background}

\subsection{Semantic Phenomena}
\label{section:semantic_phenomena}
    
        \textbf{Modality:} In this work, the focus is on any kind of modality indicating uncertainty, including modal verbs, conditionals, propositional attitudes, and negation. We see modality primarily as a signal for determining whether or not the event in question actually occurred, so that downstream applications can take this into account. We begin by discussing the typical, more specific category of modal operators.
        
        Linguistic modality communicates a speaker's \emph{attitude} towards the propositional content of their utterance. Formally, modality has been defined in terms of quantification over possible worlds \citep{kratzer2012modals}. Other definitions focus on categorising the speaker's attitude, such as epistemic necessity (\textit{That must be John.}), epistemic possibility (\textit{It might rain tomorrow.}), deontic necessity (\textit{You must go.}), and deontic possibility (\textit{You may enter.}) \citep{van2005overlap}. Sometimes a lexical trigger of modality is ambiguous between categories; English \textit{may}, for example, is ambiguous between an epistemic possibility reading (\textit{It may rain tomorrow.}) and a deontic possibility reading (\textit{You may enter.})
         
        These definitions have brought about a variety of annotation schemes in practice. \newcite{prabhakaran2012} propose five classes of modality: \textit{ability}, \textit{effort}, \textit{intention}, \textit{success}, and \textit{want}, and train a classifier on crowd-sourced annotated data. \newcite{baker2010} extend the number of modality classes to include \textit{requirement}, \textit{permission}, and \textit{belief}, and combine these with negation. \citet{penas2011overview} take a coarser, epistemic approach, asking whether events are \textit{asserted}, \textit{negated}, or \textit{speculated}, and \citet{sauri2006annotating} enrich the TimeML specification language with yet other categories (e.g. evidentiality, conditionality). 
        
        In English, modality can be expressed in a variety of ways. The modal auxiliaries (e.g. \textit{might, should, can}) are commonly used, but modality can be lexicalised in many other trigger words. Nouns (e.g. \textit{possibility}), adjectives (e.g. \textit{obligatory}), adverbs (e.g. \textit{probably}) and verbs (e.g. \textit{presume that}) can all indicate modality. In the long tail, speakers have access to vastly productive phrases that indicate their attitude. The following examples occurred naturally in the news domain \citep{zhang2013}: \textit{That's how close they were to ...}, \textit{I cannot come up with a scenario that has...}, \textit{That's based on the world wide assumption that...}.
        
        \textbf{Conditionality:} A conditional sentence is composed of a subordinate clause (which we will refer to as the antecedent) and a main clause (the consequent). The antecedent and consequent are connected by a conditional conjunction (which in English is often the word \textit{if}), as in the sentence \textit{If they attack there will be war} \citep{dancygier1999conditionals}. Conditional sentences can have a variety of semantic interpretations, but the most commonly studied, the \textit{hypothetical} conditional, expresses that the consequent (\textit{there will be war}) will hold true if the antecedent (the \textit{attack}) is satisfied \citep{athanasiadou1997conditionality}. For our purposes, the most important part of their semantics is that neither the antecedent nor the consequent are normally entailed by the sentence, so that the speaker is not committed to their truth.

    \textbf{Counterfactuality:} In the counterfactual construction a more complicated semantic relation is established between antecedent and consequent, as in the example: \textit{Had they protested, they would be content}. As with modality, this has been formalised more precisely with a possible world semantics \citep{lewis1973counterfactuals,kratzer1981partition}. With a counterfactual, the speaker communicates that in any world similar to the current one, differing only by the proposition in the antecedent, the consequent would hold true \citep{lewis1973counterfactuals}. In the above example, if the world is altered by the \textit{protest} in the antecedent, \textit{they would be content} holds true. Again, the crucial semantic information for our work is that neither the antecedent nor the consequent are entailed.
        
        \textbf{Negation} is a semantic category used to change the truth value of a proposition in order to convey that an event, situation or state does not hold \cite{horn1989}. It may be expressed explicitly using various means, most notably closed-class function words such as \textit{not, no, never, neither, nor, none} and \textit{without}, but can also be expressed lexically in open grammatical categories such as nouns (e.g. \textit{impossibility}), verbs (e.g. \textit{decline}, \textit{prevent}), and adjectives (e.g. \textit{unsuccessful}). It may also be expressed implicitly, such as with combinations of certain verb types and tenses (e.g. \textit{The polls were supposed to have closed at midnight}). In this work we consider only explicit cues of negation.

    \textbf{Propositional Attitude and Evidentiality:}
    Propositional attitude allows speakers to indicate the cognitive relations that entities bear to a proposition \citep{mckay2000propositional}. For example, in \textit{Republicans think that Trump has won}, the speaker expresses that \textit{Republicans} hold certain beliefs. In English, such reports are often made using propositional attitude verbs such as \textit{claim, warn} or \textit{believe}. Normally only the entity's thoughts regarding the event are entailed, not the event itself. Propositional attitudes are often used as markers of \emph{evidentiality} in English \citep{biber1989styles}. These are important in Question Answering. For example when answering a question using the sentence \textit{The Kremlin says protesters attacked the police} as evidence, mentioning the source (\textit{The Kremlin}) might be particularly important.

\subsection{Modality Taggers and Annotated Datasets}
A number of approaches have been proposed for the automatic tagging of modality in text. These differ in both the granularity of the classes of modality that the model tags, and the model design. 

At the lowest granularity all modality classes are collapsed into a single label. This strategy was employed in the pilot task on modality and negation detection at CLEF 2012, in which participants were asked to automatically label a set of events/states as negated, modal, neither, or both \cite{morante2012}. The submitted systems were either purely rule-based \cite{lana-serrano2012,pakray2012}, or applied rules to the output of a parser \cite{rosenberg2012}. Modality tagging has also been cast as a supervised learning task \citep{prabhakaran2012}. Performance of their classifier is reasonably strong on in-domain data (variable across 5 proposed modality classes), but out-of-domain data proves challenging. 

Due to the lack of a large, open-domain modality training dataset, we opt for a lexicon-based approach in line with that of \newcite{baker2010}. They combine a set of eight modality tags that capture \textit{factivity} with negation, to denote whether an event/state did or did not happen. They employ two strategies for tagging modal triggers and their targets: 1) string and POS-tag matching between entries in a modality lexicon and the input sentence, 2) a structure-based method which applies rules derived from the lexicon to a flattened dependency tree, inserting tags for modality triggers and targets into the sentence. 

Although there is no large, open-domain corpus in which modality is labelled, a number of small datasets exist for specific domains including biomedical text \cite{GENIA2011}, news \cite{ACEMK2017}, reviews \cite{SFU2012}, and web-crawled text comprising news, web pages, blogs and Wikipedia \cite{morante2012}.

\subsection{Event Extraction}
Since the introduction of the Open Information Extraction (OIE) task by \newcite{banko2007}, a range of open-domain information extraction systems have been proposed for the extraction of relation tuples from text. OIE systems make use of patterns, which may be hand-crafted \cite{fader2011,angeli2015} or learned through methods such as bootstrapping \cite{wu2010,mausam2012}. These patterns may be applied at the sentence level, or to semantically simplified independent clauses identified during a pre-processing step \cite{delcorro2013,angeli2015}. The majority of systems are restricted to the extraction of binary relations (i.e. relation \textit{triples} consisting of a predicate and two arguments), but systems have also been proposed for the extraction of n-ary relations \cite{akbik2012,mesquita2013}. Our system is a form of n-ary event extraction; we extract both binary and unary relations, and relations of higher valencies can be inferred by combining sets of binary relations. A comprehensive survey of OIE systems is provided by \newcite{niklaus2018}.

\section{Event Extraction System Overview}
\label{section:event_extraction_overview}

\begin{figure*}[t!]
    \small
    \centering
    \includegraphics[width=16cm]{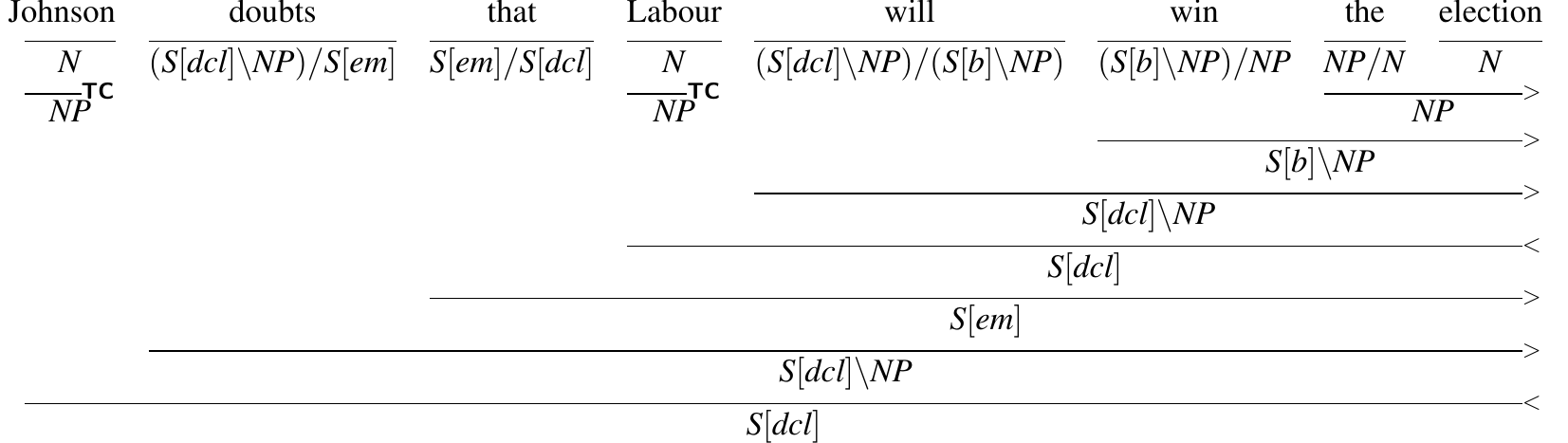}
    \caption{CCG parse tree for \textit{Johnson doubts that Labour will win the election}}
    \label{fig:CCG_parse_tree}
\end{figure*}

\begin{figure}[h]
    \centering
    \includegraphics[width=5.5cm]{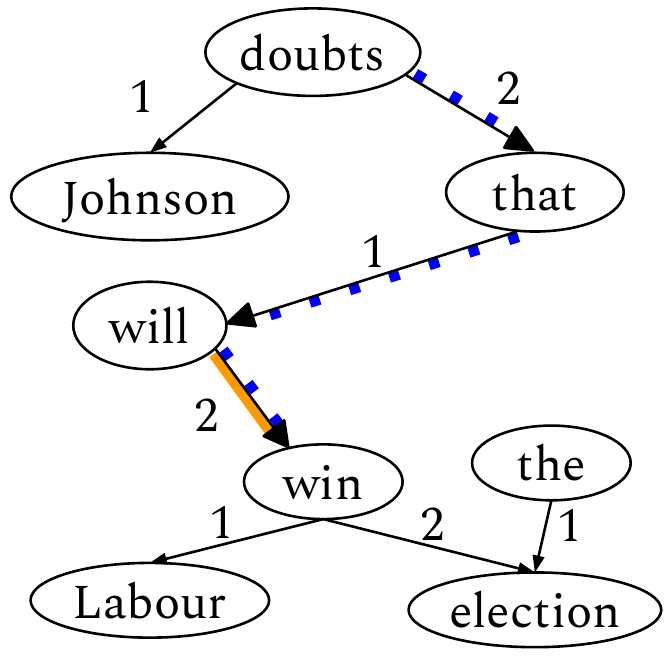}
    \caption{CCG dependency graph for \textit{Johnson doubts that Labour will win the election}; marked paths from \textit{doubts} (blue, dotted) and \textit{will} (orange, solid) to \textit{win}.}
    \label{fig:CCG_dependency_graph}
\end{figure}

Whilst many event extraction systems have been developed, none capture the wide range of modality phenomena introduced in Section~\ref{section:semantic_phenomena}.
For example, neither OpenIE nor OLLIE extract unary relations. They also fail to adequately handle all of the phenomena we are interested in, in particular counterfactuals and lexical negation. (See Section~\ref{section:extraction_system_comparison} for a comparison of our system with OpenIE and OLLIE.) We therefore construct our own event extraction system.

Our system takes as input a text document, and for each sentence outputs a set of event relations. An event relation tuple consists of a predicate and either one, or two arguments (e.g. \textbf{(The) protest}-\textit{ended}, \textbf{Angela Merkel}-\textit{addressed}-\textbf{NPD protesters}). We use a pipeline approach similar to that described by \newcite{hosseini2018}, which allows us to extract open-domain relations.

Each sentence in the document is parsed using the \textit{RotatingCCG} parser \cite{stanojevic2019} over which we construct a CCG dependency graph using a method similar to the one proposed by \citet{ccg-deps}. (See Figure~\ref{fig:CCG_dependency_graph} for an example of a dependency graph and Figure~\ref{fig:CCG_parse_tree} for the CCG parse tree from which it was extracted.) CCG dependency graphs are more expressive than standard dependency trees because they can encode long-range dependencies, coordination and reentrancies. We traverse the dependency graph, starting from verb and preposition nodes, until an argument node is reached. The traversed nodes, which are used to form the predicate strings, may include (non-auxiliary) verbs, verb particles, adjectives, and prepositions.  The CCG argument slot position, corresponding to the grammatical case of the argument (e.g. 1 for nominative, 2 for accusative), is appended to the predicate. 

Our focus is on the extraction of binary and unary relations. Binary relations may be extracted from dependency paths between two entities. Extraction of unary relations, which have only one such endpoint, poses a harder challenge \citep{szpektor2008} -- we must decide whether they are truly a unary relation, or form part of a binary relation. Therefore linguistic knowledge must be carefully applied to extract meaningful unary relations. We extract unary relations for the following cases: verbs with a single argument including intransitives (\textit{bombs exploded}) and passivised transitives (\textit{protests were held}), and copular constructions (\textit{Greta Thunberg is a climate activist}).

In addition to binary and unary relations we also extract n-ary relations which combine two binary relations via prepositional attachment. These are of the form: \textbf{arg1}-\textit{predicate-arg2-preposition}-\textbf{arg3}, and are constructed by combining the two binary relations \textbf{arg1}-\textit{predicate}-\textbf{arg2} and \textbf{arg2}-\textit{preposition}-\textbf{arg3}. For example \textbf{Protesters}-\textit{marched on}-\textbf{Parliament Square} and \textbf{Parliament Square}-\textit{in}-\textbf{London} combine to form the new relation \textbf{Protesters}-\textit{marched on Parliament Square in}-\textbf{London} (from the sentence: \textit{Protesters marched on Parliament Square in London}).

Passive predicates are mapped to active ones. Modifiers such as \textit{managed to} as in the example \textit{Boris Johnson managed to secure a Brexit deal} are also included in the predicate. As these may be rather sparse, we provide the option to also extract the relation without the modifier.

Arguments are classified as either a Named Entity (extracted by the CoreNLP \cite{manning2014} Named Entity recogniser), or a general entity (all other nouns and noun phrases). Arguments are mapped to types by linking to their Freebase \cite{bollacker2008} IDs using AIDA-Light \cite{nguyen2014}, and subsequently mapping these IDs to their fine-grained FIGER types \cite{ling2012}. For example, \textit{Angela Merkel} would be mapped to \textit{person/politician} and \textit{NPD} (Nationaldemokratische Partei Deutschland) to \textit{government/political\_party}. The type system may be leveraged to identify events belonging to specific domains, for example, to identify and track political events such as elections, debates, protests etc. and the entities involved.

\section{Lexicon}
\label{section:modal_lexicon}

Since many of the phenomena we capture involve lexical trigger items, we opt for a lexicon-based approach. Triggers identified using the lexicon can then be linked to event nodes in the CCG dependency graph. Entries in the lexicon cover modality, lexical negation, propositional attitude, and conditionality, with counterfactuality handled separately. Each entry contains the lemma, the categories that it covers, the POS-tag and an estimate of the epistemic strength that the word would normally indicate. A few examples are included in Table~\ref{table:lexicon}.

\begin{table}
\centering
\small
\begin{tabular}{@{}llcc@{}}
\toprule
Lemma & Category & POS-tag & Strength \\
\midrule
 succeed & MOD & VB & 4 \\
 shall & MOD & MD & 3 \\
 conceivably & MOD & RB & 2 \\
 impossible & MOD & JJ & 0 \\
 as long as & COND & RB & 2 \\
 concede & ATT\_SAY & VB & 4 \\
 reckon  & ATT\_THINK & VB & 2 \\
\bottomrule
\end{tabular}
\caption{Example lexicon entries}
\label{table:lexicon}
\end{table}

Our lexicon is constructed by pooling together various lexical resources. The majority of the entries derive from the modality lexicon presented by \citet{baker2010}, who use it for a similar rule-based tagging approach. Their lexicon contains just under a thousand instances, but includes multiple forms for each verb inflection. Using only infinitival forms, we add approximately 200 of the modal entries to our own lexicon.

For modelling propositional attitude, we include a list of reporting verbs found in \newcite{collins-grammar-book1990}. This added roughly another 120 phrases to the resource. The new entries were separated by attitudes expressed through speech (tag ATT\_SAY, e.g. \textit{say, state}) and attitudes of thought (tag ATT\_THINK, e.g. \textit{suspect, assume}). 

More phrases expressing uncertainty are found in a data set of news domain sentences describing conflicting events, such as a \textit{win} and a \textit{loss} \citep{guillou2020incorporating}. Such sentences often contained descriptions of events that didn't actually happen. Yet more related words were found by generating each entry's WordNet synonyms and antonyms \citep{miller1995wordnet}. We filtered and annotated these manually to obtain just under another 200 phrases, and added these to the lexicon. We also took inspiration from \citet{somasundaran2007detecting}, especially for conditionals. In aggregate, this work resulted in a resource of 530 phrases.

We also annotated each phrase with a modal category. Our lexicon contains the categories \textit{deontic}, \textit{intention} and \textit{desire}, and for the remaining phrases lists a indication of epistemic strength, with values 4 (\textit{definitely}), 3 (\textit{probably}), 2 (\textit{possibly}), 1 (\textit{probably not}) and 0 (\textit{definitely not}). The latter correspond to lexical negation. The epistemic strength values were manually annotated by the authors, and are proposed as a means to collect subsets of events, such as all events marked as \textit{probable} or higher. This phenomenon deserves more attention in future research however, as it is highly contextualised. For example, \textit{could win the lottery} should deserve a different annotation to \textit{could have breakfast}.

\section{Modality Parser}
\label{section:modal_parser}

\begin{algorithm}[t]
    \small
    \caption{Tagging Modal Events}\label{algorithm}
    \begin{algorithmic}[1]
        \Procedure{TagModalEvents}{sentence s, events e, lexicon l}
            \State $\mathcal{G}$, event\_nodes $\gets$ CCG\_dep\_parse(s, e)
            \State trigger\_nodes $\gets$ [ ]
            \For{n \textbf{in} $\mathcal{G}$}
                \If{check\_lexicon(n,l) \textbf{or} check\_cf(n,$\mathcal{G})$}
                    \State trigger\_nodes.add(n) 
                \EndIf
            \EndFor
            \For{e\_n \textbf{in} event\_nodes}
                \For{t\_n \textbf{in} trigger\_nodes}
                    \If{path\_between(e\_n, t\_n)}
                        \State e\_n $\gets$ update(e\_n,t\_n.tag)
                    \EndIf
                \EndFor
                \State e\_n.tag $\gets$ tag\_precedence(e\_n)
                \State event\_nodes.update(e\_n)
            \EndFor
            \State \textbf{return} event\_nodes
    \EndProcedure
  \end{algorithmic}
\end{algorithm}

We use the CCG-based event extraction system (Section~\ref{section:event_extraction_overview}) and the expanded modality lexicon (Section~\ref{section:modal_lexicon}) in tandem to assign modal categories to events. The procedure is described in Algorithm~\ref{algorithm}. The focus of the tagger is to identify the bulk of uncertain events: we prioritise recall over precision, so that we can expect events without a tag to have actually happened.

The event extractor produces a CCG dependency graph $\mathcal{G}$ that contains a node $n$ for each word in the sentence (line 2 of the algorithm). We then decide which of these nodes is a trigger (lines 4-7). For modality, negation, lexical negation, propositional attitude and conditionals, we tag these nodes if the node's lemma is present in the lexicon (\textit{check\_lexicon} function, line 5). The loop in the algorithm covers the simple case of single token modal triggers (such as \textit{possible}), and can be extended to multi token triggers (e.g. \textit{shoot for})\footnote{We implement this as a recursive loop over a Trie data structure.}.

Counterfactual nodes are identified separately. The \textit{check\_cf} function (line 5) finds instances of the token ``had'' that are assigned one of two indicative CCG supertags: $(((\textit{S}{\backslash}\textit{NP}){\backslash}(\textit{S}{\backslash}\textit{NP}))/(\textit{S[pt]}{\backslash}\textit{NP}))/\textit{NP}$ or $((\textit{S}/\textit{S})/(\textit{S[pt]}{\backslash}\textit{NP}))/\textit{NP}$. For example in the sentence \textit{The protesters would have been arrested, had they attacked the police}, the token ``had'' would be assigned the CCG supertag $(((\textit{S}{\backslash}\textit{NP}){\backslash}(\textit{S}{\backslash}\textit{NP}))/(\textit{S[pt]}{\backslash}\textit{NP}))/\textit{NP}$ and is therefore recognised as an instance of counterfactual had. Additionally, any instance of ``if'' that governs an instance of ``had'', is labelled as counterfactual. Upon realising that even this common counterfactual pattern was rare in the corpus, we decided not to engineer further counterfactual patterns.

We can then decide whether an event node should be tagged, by checking whether there is a path in the dependency graph from the trigger nodes to the event node (lines 9-12). Figure \ref{fig:CCG_dependency_graph} illustrates the intuition behind walking the dependency graph. The graph shows a path from both \textit{doubt} and \textit{will} to \textit{win}. This works because the existence of a path between a trigger node and an event node corresponds to the trigger node taking syntactic scope over the event node. The semantic phenomena we handle all rely heavily on this syntactic process (for example negation, see \citet{mckenna2020learning}).

A single event node may be connected to multiple triggers, so we choose the final tag on line 15. Since our primary concern is whether the event happened, we do not combine tags and instead assign a single tag based on the following order of precedence: MOD, ATT\_SAY, ATT\_THINK, COND, COUNT, LNEG, NEG. The negation categories need to be ordered last because an event that is negated and modal is still uncertain (e.g. \textit{might not play} shouldn't result in NEG\_play), but the ordering is otherwise arbitrary.

\section{Comparison with Existing Event Extraction Systems}
\label{section:extraction_system_comparison}

\begin{table*}[t]
\centering
\resizebox{\textwidth}{!}{%
\begin{tabular}{@{}lll@{}}
\toprule
\textsc{MoNTEE} & OpenIE & OLLIE \\
\midrule
\multicolumn{3}{@{}l@{}}{\textbf{The guerrillas are ready to talk with the Soviets, if Moscow is willing.}} \\
\textsc{mod}\_(guerrillas; talk; Soviets) & (guerrillas; are; ready) & (Moscow; is; willing) \\
\textsc{cond}\_(Moscow; be willing)  & (guerrillas; talk with; Soviets) & \\
& (guerrillas; talk; if Moscow is willing) & \\
& (guerrillas; talk; willing) & \\
& (Moscow; is; if Moscow is willing) & \\
& (Moscow; is; willing) & \\
\midrule
\multicolumn{3}{@{}l@{}}{\textbf{Had Trump won the election, Cummings would still be in Downing Street.}} \\
\textsc{count}\_(Trump; win; election) & (Trump; Had Trump won; election) & (Trump; Had won; the election) \\
\textsc{mod}\_(Cummings; be in; D.St.) & (Cummings; would; would still be in D.St.) & (Cummings; would still be in; D.St.) \\
\midrule
\multicolumn{3}{@{}l@{}}{\textbf{Protesters did not attack the Police.}} \\
\textsc{neg}\_(Protesters; attack; police) & $\varnothing$ & (Protesters; did not attack; the police) \\
\midrule
\multicolumn{3}{@{}l@{}}{\textbf{Parliament failed to investigate the Kremlin.}} \\
(Parliament; failed to investigate; Kremlin) & (Parliament; investigate; Kremlin) & (Parliament; failed to investigate; the Kremlin) \\
\textsc{lneg}\_(Parliament.; investigate; Kremlin) & & (Parliament; to investigate; the Kremlin) \\
\midrule
\multicolumn{3}{@{}l@{}}{\textbf{Ed Miliband says the government betrayed Yorkshire.}} \\
\textsc{att\_say}\_(government; betray; Yorkshire) & $\varnothing$ & (the government; betrayed; Yorkshire) \\
(Ed-Miliband; say)  &  & [attrib=Ed Miliband says] \\
\bottomrule
\end{tabular}}
\caption{Comparison of \textsc{MoNTEE} with OpenIE and OLLIE}
\label{tab:system_comparison}
\end{table*}

We highlight the capabilities of our system on five example sentences, comparing with two existing event extraction systems: OpenIE \cite{angeli2015} and OLLIE \cite{mausam2012}. Note that this is not intended as a conclusive evaluation of systems, but rather as a high-level overview of the phenomena captured by each of the systems. See Table~\ref{tab:system_comparison} for a comparison of the relations extracted by \textsc{MoNTEE}, OpenIE and OLLIE. The examples are all naturally occurring sentences from the news domain, obtained by a web search targeted to the modality categories discussed in this paper. To enable a fair comparison, we focus on the extraction of binary relations, as neither OpenIE nor OLLIE was designed to extract unary relations.

Whilst Stanford OpenIE \cite{angeli2015}, OLLIE \cite{mausam2012}, and OLLIE's predecessor \textsc{ReVerb} \cite{fader2011} may be used to extract binary relations for events, they do not explicitly mark events for modality or negation. Stanford OpenIE \cite{angeli2015} typically includes modals as part of the predicate (for example: (Protesters; may have attacked; police)), but ignores the other categories of linguistic modality described in Section~\ref{section:modal_parser}. In particular it does not extract relations for sentences involving negation or propositional attitude, omits lexical negations, and is easily confused by sentences involving conditionals or counterfactuals.

OLLIE \cite{mausam2012} handles the phenomena in more detail. It identifies conditionals by detecting markers such as ``if'' and ``when'', and labels the \textit{enabling condition} for extracted relations that are governed by a conditional\footnote{The labelling of \textit{conditional} is not applied in the first example in Table~\ref{tab:system_comparison} as no relation is extracted for the consequent.}. It typically includes modals and negation as part of the predicate, and captures propositional attitude in its handling of attribution (e.g. \textit{Ed Miliband says...}). Like OpenIE, OLLIE is not designed to handle counterfactuals. In terms of lexical negations, OLLIE extracts the predicate both with and without the negation cue, which is undesirable if the downstream NLP application needs to be able to distinguish between events that took place and those that did not.

\section{Evaluating System Performance}
\label{section:intrinsic_evaluation}

In the absence of a pre-existing open-domain evaluation dataset that closely matches the task we are interested in, we conduct an intrinsic evaluation of our modality-aware event extraction system. We measure performance on a set of 100 extracted event relations with manually annotated labels denoting the degree of certainty (happened, didn't happen, uncertain). An event relation consists of a predicate plus argument pair (e.g. (Protesters; attack; police)). Note that we exclude both OLLIE and OpenIE from this evaluation as neither system is designed to handle the complete set of modality or negation phenomena we are interested in (c.f. Section~\ref{section:extraction_system_comparison}).

We filtered the articles in the NewsSpike corpus \cite{zhang2013} to obtain those where at least 20\% of the event relations are tagged (to guarantee a reasonably dense distribution of modality). We then randomly selected five articles and processed them using our system to extract event relations. From these articles we selected 100 event relations\footnote{We excluded those event relations for which the predicate contains only a preposition as these have little meaning unless they form part of a high-order n-ary relation.}. At the sentence-level we ensured that we include only one event relation for each predicate node in the dependency graph, since all event relations with the same predicate node will be assigned the same modality. 

The set of 100 event relations was manually annotated by two of the authors of this paper, one native English speaker and one fluent speaker. For each event relation, we asked the annotators to answer the question \textit{Does the text entail that the event definitely happens?} using the following labels: the event happened (2), is uncertain (1), didn't happen (0). Inter-annotator agreement over the set of 100 event relations was measured using Cohen's Kappa \cite{Cohen1960}. The agreement score was 0.77, indicating \textit{substantial agreement}, and the annotations differed for only 16 examples. Following the initial annotation task, the two annotators resolved the disagreements, which resulted in the gold standard test set.

To evaluate our system, we mapped system-assigned modal and negation tags to the set of certainty labels, with LNEG and NEG tags mapped to 0 (didn't happen), empty tags mapped to 2 (happened), and all other tags mapped to 1 (uncertain). In Table~\ref{tab:intrinsic_results} we report the micro- and macro-averaged precision, recall and F1 scores. As the number of event relations per modality tag category is too small for a meaningful error analysis over types, we provide aggregated scores. The distribution of certainty labels is also uneven, with few negations marked in the gold standard. We therefore take the micro-averaged F1 score of 0.81 to be the definitive result.

\begin{table}
\small
\centering
\begin{tabular}{lccc}
\toprule
& Precision & Recall & F1 \\
\midrule
Micro-average & 0.81 & 0.81 & 0.81 \\
\midrule
Macro-average & 0.72 & 0.88 & 0.76 \\
\bottomrule
\end{tabular}
\caption{Intrinsic evaluation results}
\label{tab:intrinsic_results}
\end{table}

We performed an error analysis of the 17 errors made by our system on the test set of 100 event relations. Parsing was a common issue, with five errors attributed to general parsing mistakes, and five errors due to missing dependency links between reporting verbs and events in quoted text (e.g. \textit{``Police were attacked'', they said}). Two mistakes were due to human error, as the annotators also missed these reporting verbs in longer sentences. Then, three errors arose from issues with the lexicon. Two of these stemmed from lack of coverage: our lexicon does not handle temporal displacement, as in \textit{We won't act \textbf{until} the white house gives more information}. The other was caused by incorrect application of a lexical entry, which would need to be disambiguated using context. Finally, two errors could also have been avoided by handling linguistic \textit{aspect}, as in \textit{they began the process to...}. Future research could thus focus on expanding the lexicon by these final categories of displacement, and taking context into account when linking a word to the lexicon.

\section{Corpus Analysis}

We conducted a corpus analysis of extracted relations over the NewsSpike corpus \cite{zhang2013}. NewsSpike contains approximately 540K multi-source news articles (approximately 20M sentences) collected over a period of six weeks. We report on the distributions of tagged phenomena over the set of binary relations\footnote{The corpus study of unary relations is left for future work} extracted from news articles in the complete corpus (general domain), and for the subsets of articles related to the politics and sports domains.

The NewsSpike corpus does not include topic or domain information in the article-level metadata. Therefore to identify articles belonging to the politics and sports domains we leveraged the named entity linker AIDA-Light \cite{nguyen2014} and the FIGER  type system \cite{ling2012}. We first identified the set of fine-grained FIGER types related to each sub-domain, and then obtained the set of entities belonging to each type. Next we used the output of AIDA-Light to identify the set of articles for which more than 40\% of the entities found by the linker belonged to the politics domain, with at least two political entities. We repeated this process for the sports domain, with a lowered threshold of 25\%, as the sports topic is less likely to overlap with other topics.

The distribution of relation tags over the general, politics, and sports domains is shown in Table~\ref{tab:modal_stat_summary_by_domain}. For the politics domain just over 25\% of the extracted relations are tagged by the modality parser, which is more than for the sports or general domains. In particular, modals are more prevalent. This suggests that whilst it is important to identify modality in the general news domain, it is particularly important in the politics domain.

The top ten most frequent trigger words found in the general domain are: the propositional attitude trigger \textit{say}, the modal triggers \textit{will}, \textit{would}, \textit{can}, \textit{could}, \textit{may}, \textit{should}, \textit{want} and \textit{have to}, and the conditional trigger \textit{if}. The same top ten are also observed for the politics domain (with different frequencies), and for the sports domain the propositional attitude trigger \textit{think} replaces \textit{want}. The similarity of these lists is perhaps not surprising as all three domains belong to the news genre.

\begin{table}
\small
\centering
\begin{tabular}{@{}lrrr@{}}
\toprule
& General & Politics & Sports \\
\midrule
Articles & 532,651 & 58,521 & 196,098 \\
Sentences & 20,683,584 & 2,280,312 & 8,056,704 \\
\midrule
Relations & 96,774,467 & 11,265,585 & 37,936,677 \\
\midrule
\multicolumn{4}{c}{Distribution of tags (percentage of all relations)} \\ [0.15cm]
$\varnothing$ & 77.83& 74.78 & 78.75 \\
Tag & 22.17 & 25.22 & 21.25 \\
\midrule
\multicolumn{4}{c}{Distribution of \textit{types} of tag (percentage of tagged relations)} \\ [0.15cm]
Modal & 64.59 & 66.04 & 65.10 \\
ATT\_say & 21.54 & 21.28 & 19.94 \\
ATT\_think & 2.22 & 1.72 & 2.32 \\
Conditional & 4.03 & 4.09 & 3.99 \\
Counterfactual & 0.17 & 0.19 & 0.19 \\
Negation & 6.86 & 6.00 & 7.79 \\
Lexical Negation & 0.58 & 0.67 & 0.67 \\
\bottomrule
\end{tabular}
\caption{Relation tagging summary by news domain}
\label{tab:modal_stat_summary_by_domain}
\end{table}

\section{Future Work}

An obvious limitation of our approach is that it does not take into account the context in which events and trigger words occur. Modality is a context-dependent phenomenon, so using the sentential context would improve accuracy. For example, the word \textit{unbelievable} is ambiguous between an \textit{unlikely} and an \textit{amazing, and happened} reading. Relatedly, our concept of epistemic strength is highly context-sensitive, and requires further development. A promising avenue is to develop a pre-training procedure for a modality-aware contextualised language model \cite{devlin2019, zhou2020temporal}. We plan to use our modal lexicon to identify sentences with modality triggers. We will then gather human annotations of the certainty that each event happened, and use this annotated data to train a modality-aware language model able to classify event uncertainty. Such a system might eventually even tackle the long-tail of modal examples mentioned in Section~\ref{section:semantic_phenomena}.

We will also investigate the application of zero shot and few shot learning to the problem of detecting modality and negation. This could provide a way to leverage a large pre-trained language model together with a small annotated corpus.

Our system was developed for English, but work is already underway to develop event extraction systems for other languages including German and Chinese. Extending to other languages would allow us to apply our methods to multilingual and cross-lingual NLP tasks. Finally, most CCG parsers, including the one used in this work, are trained on English CCGbank \citep{ccgbank}. This makes them perform well on news text, but accuracy suffers on out-of-domain sentences, primarily those involving questions.
The results could be improved by retraining the parser on the CCG annotated questions dataset \citep{ccg-questions,yoshikawa-2019}, allowing us to apply our system to the task of open-domain Question Answering in an extrinsic evaluation.

\section{Conclusion}

We have presented \textsc{MoNTEE}, a modality-aware event extraction system that can distinguish between events that took place, did not take place, and for which there is a degree of uncertainty. Being able to make such distinctions is crucial for many downstream NLP applications, including Knowledge Graph construction and Question Answering. Our parser performs strongly on an intrinsic evaluation of examples from the politics domain and our corpus analysis supports our claim that modality is an important phenomenon to handle in this domain.

\section*{Acknowledgments}

This work was funded by the ERC H2020 Advanced Fellowship GA 742137 SEMANTAX and a grant from The University of Edinburgh and Huawei Technologies.

The authors would like to thank Mark Johnson, Ian Wood, and Mohammad Javad Hosseini for helpful discussions, and the reviewers for their valuable feedback. 

\bibliographystyle{acl_natbib}
\bibliography{acl2021}

\end{document}